# Problems With "Programmable self-assembly in a thousand-robot swarm"

Author: Muaz A. Niazi*[‡]

**Affiliations:**

Department of Computer Science,

COMSATS Institute of IT,
Islamabad, Pakistan

*Correspondence to: muaz.niazi@ieee.org,

**Abstract**: Rubenstein et al. present an interesting system of programmable self-assembled structure formation using *1000* Kilobot robots. The paper claims to advance work in artificial swarms similar to capabilities of natural systems besides being highly robust. However, the system lacks in terms of matching motility and complex shapes with holes, thereby limiting practical similarity to self-assembly in living systems.[1]

**Main Text:**

Self-assembly is an exciting idea often found in living systems. Whitesides and Grzybowski present an in-depth review of self-assembly classifying self-assembly into four types: static, dynamic, template and biological (*1*). The basic ideas of self-assembly by design or imposed self-assembly may have origins dating as far back as the 1940's in the form of concepts of self-replicating machines as well as cellular automata by computing pioneers Von Neumann and Burks (*2*, *3*). More recently, other variations of programmable self-assembly have been presented as part of John Holland's work on Genetic algorithms which use an evolution of populations of 'chromosomes' (*4*). Still another direction of programmable self-assembly is a set of ideas of genetic programming with a general focus on the evolution of programs to user requirements (*5*). While 3-D printing technology may soon lead to large-scale artificial self-replicating and self-assembly systems, Rubestein et al. (*6*) present a novel system with a large number of robots capable of self-assembly forming different shapes.

---

[1] The comment was submitted to both the authors as well as the Science magazine but no response to the technical problems with the manuscript was received from either the journal or the authors.

To summarize, the system presented in the paper self-assembles using some of the following key concepts:

A. Formation of a cluster of seed robots manually by the human user for establishing a gradient system.

B. Establishment of a gradient by calculation and communication between individual robots.

C. Use of the gradient by Kilobots for "edge following" while keeping distance from other nodes, moving or otherwise. Essentially, robots move along the periphery of the existing aggregate of robots to gradually move to a position which is acceptable in relation to the figure input by the user to all Kilobots.

While the work by the authors is certainly commendable, science is based on a continuous process of learning, adaptation and improvement, something which is also the norm of humans and other prevailing species around the globe, both at the individual level as well as collectively as a species or a civilization. As such, here I would like to humbly note technical reservations to certain aspects of the system based on which, it is clear that there is not ample evidence to support the results of matching real-world self-assembly systems as discussed next:

1. The presented system is limited to catering only for solid shapes with "no holes" understandably because the robotic movement is based on following the edge by keeping a distance from existing robotic aggregate as can also be observed in the supplementary material/videos. This is a limitation of the gradient-based approach. Perhaps if the requirement of a gradient is removed, this problem could have been avoided altogether though it might require a fundamental re-design of the self-assembly system.

2. Additionally, in practice, the actual algorithm only engages the robots at the periphery to form the pattern. All robots start in an aggregated form rather than being placed at random places, which resultantly limits the utility or correlation with real-world systems.

3. The system requires manual positioning of seed robots by human operators; again a somewhat unrealistic requirement limited the match with real-world self-assembly systems and thereby making the system only partially self-assembled. It could however have been corrected with an initial execution of leader election algorithm such as presented in (*7*), assuming if the Kilobots were given wireless communication capability instead of the existing communication mechanism (Discussed later).

4. Currently the Kilobots move based on a comparatively higher value of the gradient—a limitation because it implies that the robots only work if they have neighbors. If there are no neighbors, there can be no communication of a gradient value and hence the algorithm might fail.

5. Each Kilobot is capable of communication using short-range infra-red reflection up to a distance of *10* cm. This communication mechanism, although ingenious, is based on reflection from the base table surface for robots thereby limiting applications based on non-reflective or rough surfaces besides being unlike the communication mechanisms of real-world swarms (*8*) (which are often based on optical, aural, sensory or pheromone-based sensing)

6. While the motility of Kilobots is based on an unconventional, albeit novel concept of motion using vibrating motors, the mechanism makes individual robot movement slow and cumbersome, not in line with the scale of motility in real-world cells such as E. coli and M. jannaschii – one being capable of moving at the rate of *20* and the other up to *500* body lengths per second (*9*). The resultant problems are evident in the simulation with some simulations taking 11+ hours.

7. Finally, the system also requires manual removal of leftover robots after the completion of the shape. In other words, the algorithm again requires human user intervention at two additional levels: first to visually evaluate the shape for completion and secondly to manually remove the extra robots no longer needed by the system (detailed in the supplementary material).

To summarize, while the presented system is a novel advancement in the state of the art, , there is currently room for improvement in the existing system for making it more realistic in terms of self-assembly and robustness to match real-world self-assembly systems unlike concluded in the paper.